\documentclass[sigconf]{acmart}

\AtBeginDocument{%
  }

\copyrightyear{2025}
\acmYear{2025}
\acmDOI{XXXXXXX.XXXXXXX}
\acmConference[Preprint]{Preprint}{Febuary 07, 2025}{Redmond, USA}

\usepackage{xspace}
\usepackage{subcaption}
\usepackage{graphicx}
\usepackage{booktabs}

\usepackage[textsize=tiny, disable]{todonotes}

\newcommand{\alerting}{\textsc{Alerting}\xspace}
\newcommand{\hil}{\textsc{Human in the Loop}\xspace}

\newcommand{\streaming}{\textsc{Streaming}\xspace}
\newcommand{\populations}{\textsc{Populations}\xspace}
\newcommand{\conditionalanomalies}{\textsc{Conditional Anomalies}\xspace}
\newcommand{\applicationspecific}{\textsc{Application Specific}\xspace}

\newcommand{\pointprocesses}{\textsc{Point Processes}\xspace}
\newcommand{\windowsize}{\textsc{Periocity}\xspace}
\newcommand{\thresholdsetting}{\textsc{Threshold Setting}\xspace}
\newcommand{\explainable}{\textsc{Explainability}\xspace}

\newcommand{\xtt}{\mathbf{x}_{<t}}
\newcommand{\ytt}{\mathbf{y}_{<t}}
\newcommand{\ztt}{\mathbf{z}_{<t}}

\begin{document}

\title{Open Challenges in Time Series Anomaly Detection: An Industry Perspective}

\author{Andreas C. M\"uller}
\email{amueller@microsoft.com}
\affiliation{%
  \institution{Microsoft}
  \country{USA}
}








\renewcommand{\shortauthors}{M\"uller}

\begin{abstract}
Current research in time-series anomaly detection is using definitions that miss critical aspects of how anomaly detection is commonly used in practice. We list several areas that are of practical relevance and that we believe are either under-investigated or missing entirely from the current discourse. Based on an investigation of systems deployed in a cloud environment, we motivate the areas of streaming algorithms, human-in-the-loop scenarios, point processes, conditional anomalies and populations analysis of time series. This paper serves as a motivation and call for action, including opportunities for theoretical and applied research, as well as for building new dataset and benchmarks.
\end{abstract}



\maketitle

\todo{For each section, what does current literature do, what do we need?}
\todo{Should A produce a score or a label?}

\section{Introduction}
Recent years have seen a renewed academic interest in time series anomaly detection (TAD), with several new methods being described, as well as broad discourse on metrics, benchmark datasets and categorization of anomaly types~\citep{wu2021current, lai2021revisiting, schmidl2022anomaly, paparrizos2022tsb, liu2024elephant, liu2024time, sarfrazposition, mejri2024unsupervised, gungor2025robust}.
In this paper, we argue that while laudable and highly relevant, these efforts miss some fundamental aspects of time series anomaly detection in practical applications. This work explains these missing aspects, and why they are important for practical applications. Our perspective is informed by working with both internal and external customers at a cloud provider.
In this paper we will argue for the following two core tenets that we believe should drive future research in time series anomaly detection:
\begin{description}
    \item[\alerting] The main applications of time series anomaly detection are alerting and mitigation, i.e. either letting  a human know that an anomalous event happened, or automatically reacting to an anomalous event. 
    \item[\applicationspecific] Whether a certain event is considered an anomaly in practice is not intrinsic to the data, but is application specific, i.e. an event that would be considered anomalous in one application could be considered normal in another.
\end{description}
We will elaborate on these two tenets in Section~\ref{sec:alerting} and Section~\ref{sec:contextual}. The first tenet is an empirical observation from applications, which can also be found in the applied TAD literature, which is extremely biased towards alerting scenarios, including Space Flight~\citep{hundman2018detecting}, cloud operations~\citep{laptev2015generic, ren2019time, vallis2014novel, james2016leveraging}, building maintenance~\citep{yu2016advances}, smart grids~\citep{zhang2021time}, 
financial fraud~\citep{hilal2022financial} and cyber security~\citep{ten2011anomaly}. In fact, it is hard to find practical applications of anomaly detection that do not require timely alerting.
The second tenet is an assumption about the nature of anomalies relevant in practice%
. It is also motivated by the difficulties in evaluating current algorithms, and the absence of broadly applicable solutions, see the discussion in Section~\ref{sec:surveys}. We discuss alternative viewpoints in Section~\ref{sec:alternativeviews}.
From these two tenets, we derive the following requirements:
\begin{description}
    \item[\streaming] Alerting requires streaming algorithms to efficiently update on newly arriving data (a computational optimization). More importantly, it also requires streaming evaluation, i.e. using only previously seen data for scoring a given datapoint, a fundamental requirement in algorithm comparison.
    \item[\populations] Alerting is usually done on a population of semi-homogeneous time-series, i.e. monitoring the room temperature in all rooms in a building, or monitoring the traffic or disk usage on the nodes of a cluster, or app usage per user for a mobile app. From these populations, depending on the scenario, different levels of alerting, say on the whole population, on individuals, or on sub-populations could be desired.
    \item[\conditionalanomalies] Alerting often requires external knowledge to determine whether a particular behavior is expected or anomalous. For example, treating sales as a purely univariate TAD problem, the 4th of July might show zero sales in a store, which would be interpreted as anomalous. Most domain experts would not consider this an anomaly, however, since the behavior could be easily explained away by the fact that it is a holiday, and the store is closed.
    \item[\hil] If anomalies are not intrinsic to the data, but application specific, it's essential to have a human in the loop, and detect relevant anomalies based on human feedback.
    \item[\explainable] In any alerting system, users are likely to ask two ``why'' question for any given alert. The first is ``Why was an alert raised?'', the second is ``What caused the alert?'', corresponding to explainability and root cause analysis (RCA).
\end{description}
In addition to these requirements that follow from our core tenets, we have also found several requirements that are related to the low-level processing of signals. These requirements are critical to applying TAD, but are not usually discussed in the literature.
\begin{description}
    \item[\pointprocesses] Data is often presented as point processes (i.e. not observed at regular intervals) while TAD algorithms require timeseries data.
    \item[\windowsize] Most TAD algorithms require a window size, which is often determined automatically from data. The method used for determining the window size are not standardized, often not even discussed, different implementation yield widely varying results, and the impact of these on TAD algorithms is not well-understood.
    \item[\thresholdsetting] Threshold setting is a critical component of any real-world TAD system, though usually treated as a separate issue. Given the usual unsupervised nature of the problem, even with human intervention, it's unclear how to set a threshold effectively.
\end{description}
In the following sections, we will argue why these are requirements of real-world deployments of TAD, and why current research does not sufficiently address these requirements. We provide an motivating example based on a real use-case in Section~\ref{sec:fridge}. We discuss alerting requirements in Section~\ref{sec:alerting} and propose explicit formulations of the TAD setup that satisfy some of these requirements, in particular cohort-level TAD and Conditional TAD in Section~\ref{sec:side_information}. We discuss the online setting with human feedback in Section~\ref{sec:contextual} and propose a natural formulation resembling a censored bandit problem. We discuss pre-processing and normalization requirements in Section~\ref{sec:signalprocessing} and end with discussing several counter-arguments to our tenets and requirements in Section~\ref{sec:alternativeviews}.

%



\subsection{Related Work}
\subsubsection{Surveys of Time Series Anomaly Detection, Metrics and Definitions}\label{sec:surveys}
In recent years, there has been a wealth of papers reviewing, classifying and benchmarking TAD algorithm and datasets, providing a wide array of insights into the characteristics of algorithms and dataset.
This wave of reviews was started by the seminal work \cite{wu2021current} that proclaimed ``Current time series anomaly detection benchmarks are flawed and are creating the illusion of progress'' and showed how simple heuristics can solve most popular benchmark datasets used in the literature at that point.
\citet{lai2021revisiting} provide a qualitative definition of different types of anomalies, and introduce formal characterizations of different types of anomalies in an elegant but limited framework.
Given the criticism of existing benchmarks, \citet{schmidl2022anomaly} and \citet{paparrizos2022tsb} concurrently proposed more expansive benchmarks and provided an extensive evaluation of existing methods, often finding classical and relatively simple methods to work well. \citet{schmidl2022anomaly} also provides a taxonomy and survey of current methods, and evaluates their performance given several synthetic anomaly scenarios. \citet{sarfrazposition} showed that very simple baselines can not only match, but even outperform complex deep learning solutions, and that the commonly used adjusted F1 metric is flawed. \citet{kim2022towards} further disseminate this metric and propose alternatives, which is further built upon by \citet{paparrizos2022volume}.
\citet{liu2024elephant} updated the benchmark provided by \citet{paparrizos2022tsb} and used human annotations and consistency rules to curate a diverse subset of the original benchmark that contains consistent labels. 
\citet{liu2024time} provides another taxonomy of recently proposed algorithms. \citet{mejri2024unsupervised} provide a meta-survey, comparing recent benchmarking papers, and introduce a novel experimental protocol.
\citet{sorbo2024navigating} provide a thorough review of different evaluation metrics and their effects on algorithm ranking.

We want to highlight some common themes in these works; first, there is no accepted definition of anomaly, though several definitions have been proposed. However, most definitions are rather qualitative and not formally specified, with the exception of \citet{lai2021revisiting}. At the same time, datasets have been criticized either for being overly simple, unrealistic, or inconsistently annotated. We posit that these issues are interrelated, as the vague definition of the problem makes it difficult to define what an appropriate benchmark should look like. Several synthetic benchmarks have been proposed~\citep{schmidl2022anomaly, lai2021revisiting} but without consistent definitions, it's unclear whether the assumptions made there are realistic for real-world scenarios.
A possible explanation for the continued disagreement about evaluation and datasets could be our tenet \applicationspecific, which states that without additional information or assumptions, the problem is ill-specified and potentially unsolvable.

We also want to point out that despite this flood of benchmarks and surveys, drawing conclusions is still extremely difficult~\citep{sorbo2024navigating}; two of the most comprehensive benchmarks, \citet{schmidl2022anomaly} and \citet{liu2024elephant} come to very different conclusions: \citet{schmidl2022anomaly} finds LSTM-AD to be one of the highest ranked algorithms, while \citet{liu2024elephant} finds it to be ranked in the middle, outperformed by simple methods like $k$-means (and \citet{paparrizos2022tsb} ranks is near the bottom), while \citet{liu2024elephant} showed good success with MCD~\citep{rousseeuw1999fast}, which was excluded from the analysis in \citet{schmidl2022anomaly} for common failure.

\subsubsection{Streaming}\label{sec:related_streaming}
Despite the extreme prevalence of streaming applications of TAD, we are aware of only two recent works addressing the streaming setting, \citet{ren2019time}, who propose a simple score based on spectral residuals, which are then optionally re-calibrated using a CNN trained on synthetic data, and \citet{boniol2021sand}, who propose a clustering based algorithm using shape-based distances. \citet{schmidl2022anomaly} discuss the ability to operate on streams in their taxonomy, but do not evaluate the streaming setting. Unfortunately, \citet{ren2019time} and \citet{boniol2021sand} only provide limited benchmarks, both in terms of datasets considered and in terms of baseline algorithms.

\subsubsection{Human In The Loop}
Somewhat surprisingly, there is little work on using human annotations in TAD in the academic literature. \citet{liu2015opprentice} describe a system that is used in a production setting and uses human annotations. However, the system is purely supervised, and requires a significant amount of human annotation. Evaluation is limited to the KPI dataset; however, the work includes many aspects of TAD that are extremely relevant in practical applications, but have been neglected by the academic community. A more recent approach focusing on the user interface, is described in \citet{deng2024reliable}.

\subsubsection{Root Cause Analysis}
Traditionally, root cause analysis (RCA), i.e. finding the cause of an anomaly, has been treated as a separate problem from anomaly detection~\citep{soldani2022anomaly}. However, \citet{yang2022causal} proposed an integrated causal framework for TAD and RCA, which allows for addressing \conditionalanomalies. This work is likely motivated by real applications, given the industry affiliation; however it has received little attention. While there is a large array of work in the operations and BI community, as reviewed by \citet{soldani2022anomaly}, the area seems not well studied within the ML community.

\subsubsection{Signal processing for TAD}
There are two critical signal processing components involved in most TAD applications: resampling or aggregating to a time series, and detecting periodicity lengths. While resampling signals is a well-studied problem in signal processing, the usual goal is to remove artifacts and obtain smooth signals~\citep{oppenheim1999discrete} which at odds with the goal of anomaly detection, that usually seeks extreme and unusual values. To the best of our knowledge, this interaction has not been studied.
Finding the dominant periodicity of a signal has been studied in the data mining community~\citep{vlachos2005periodicity} and the database community~\citep{elfeky2005periodicity, wen2021robustperiod}.
However, despite many TAD algorithms requiring a window length, the impact of this choice has not been systematically studied in any of the surveys on TAD~\ref{sec:surveys}, and this length has been either manually specified, or set using simple heuristics. Recently, \cite{ermshaus2023window} evaluated several methods for window size selection; however, the benchmark considers only a small subset of the algorithms considered in the other review articles, and uses the discredited adjusted F1-score.

\subsection{Definition and Notation}
The notion of anomaly in a time series is difficult to formalize, as discussed in Section~\ref{sec:surveys}. Without a clear mathematical characterization of what an anomaly is, however, formulating different problem settings and requirements if difficult, if not impossible.
\todo{appendix with definitions}
We attempt to introduce a relatively general formal notation here to have a more precise language to describe problem settings, not for the purpose of a theoretical analysis of the assumptions, nor as a practical tool.
We assume that any time-series under consideration $\mathbf{x} = (x_t)_{t=1,\dots,n}$ is drawn from a distribution 
\begin{align*}\label{eq:distribution_assumption}
    \bar{p}(x_t|\xtt) = &(1-\epsilon) p(x_t|\xtt) + \epsilon q(x_t|\xtt)
\end{align*}
where here and in the following, \[\xtt = (x_1, \dots, x_{t-1}).\]
Here $\epsilon$ is the probability of any point $x_t$ being an anomaly, $q$ is an arbitrary distribution, and $p$ conforms to a Markov process of unknown order $k\ll n$, i.e.
\begin{equation*}
    p(x_t| \xtt) = p(x_t|x_{t-k}, \dots x_{t-1}),
\end{equation*}

The intuition is that $\bar{p}$ is a mixture of a regular distribution $p$ and a noise distribution $q$. We call $x_t$ an anomaly when $p(x_t|\xtt) < \eta$ for some threshold $\eta >0$, i.e. if $x_t$ is unlikely under the noise-free distribution.
The definition of $\bar{p}$ only captures \emph{point anomalies} as described in \citet{lai2021revisiting} as each point is independently drawn from regular or noise distribution; modeling pattern anomalies is possible by making $\epsilon$ a function of $\xtt$; however, for the arguments in this paper, limiting ourselves to point anomalies is sufficient, even though pattern anomalies are also extremely relevant in practice.

\section{An Illustrative Example: Temperature Sensors}\label{sec:fridge}
We start our discussion of our requirements with a representative example of a slightly simplified real-world use-case in which TAD was applied to illustrate these requirement in the context of an application.

Consider a nation-wide retail chain selling fresh groceries. This requires a network of warehouses with refrigerators and cooling trucks. Each cooling unit has a thermometer and is monitored for being in the allowable range given its contents. This means, depending on the scale of the company, hundreds, thousands or tens of thousands of time series that are of the same nature. If a certain unit is above the allowable temperature for a certain time, the food can no longer be sold. Any abnormality in the data that could indicate an imminent incident is therefore business critical. Clearly, \streaming is required, as notification of anomalies should be as fast as possible, and future data is not available. Also, new data in the form of current temperature measurements needs to be constantly incorporated.
When using the data for alerting, there are several possible consumers of the alert. Most importantly, there might be an operator or team that can investigate failures. Constantly monitoring tens of thousands of series is impractical, and alerting on individual series might be overwhelming. Often multiple anomalies are correlated, for example if the cooling units are co-located. If a truck with several units crashes, several might open and change their temperature, or they could short-circuit and no longer send signals. Ideally, the system would alert the operator about a subpopulation of anomalies associated with this truck. Depending on the scenario, this could either fall in \populations, if the assignment of units to locations is fixed (i.e. a unit is always installed in the same truck, warehouse, or store, so the assignment independent of time), or into \conditionalanomalies if the units travel between trucks, warehouses and stores, leading to a time series of locations associated with each unit.
In either case, alerting should likely be at the most general level possible, say, alerting about a regional power-outage as something happening in a region, not on hundreds of individual units.

\begin{figure*}[ht]
    \centering

        \includegraphics[width=\textwidth]{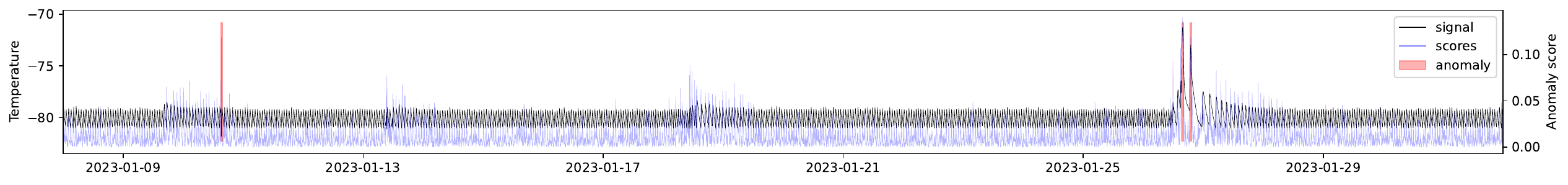}
    \caption{Temperature data for an ultra-low temperature freezer (Unit Haier 810545 from \citet{huang2023labelled}. Anomaly scores are computed with spectral residual~\citep{ren2019time} and thresholded at the 99.9th percentile. The two detected anomalies have very different patterns and real-life consequences.}
    \label{fig:temperatureanomalies}
\end{figure*}

While the main consumer of anomaly reports on the cooling units might be an operations team, another usage is device  maintenance. For example, a temperature sensor that shows extreme spikes for short periods of time is likely malfunctioning, and needs to be replaced. This is of no consequence to the logistics operations, since short high temperature spikes are physically implausible, and therefore measurement errors when it comes to the actual state of the cooling unit. However, from the perspective of a maintenance team, this malfunctioning sensor clearly needs attention and might need replacement. Here, the notion of anomaly does not only depend on the signal, but also on the downstream usage, exhibiting \applicationspecific. A real-live example of this is shown in Figure~\ref{fig:temperatureanomalies} using data of ultra-low temperature freezers, commonly used for pharmaceutical applications~\cite{huang2023labelled}. The first anomaly marked in the figure shows an extreme outlier: a dip below the standard temperature of $-81^\circ C$. The second anomaly consists of two large spikes of temperatures of between $-72^\circ C$ and $-74^\circ C$. From a statistical point of view, both are significant anomalies, even using very simple measures. The low temperature is cause by a temporary power outage, the high temperature by the door to the unit being left open. From an operational perspective, only the second event is likely to be relevant.

\section{Alerting Requirements}\label{sec:alerting}
\subsection{Streaming}\label{sec:streaming}
As discussed in Section~\ref{sec:related_streaming}, there have been several efforts in streaming anomaly detection \citep{boniol2021sand, ren2019time}. However, even though some work discuss the streaming capabilities of certain algorithms~\citep{schmidl2022anomaly}, there exist few actual implementations of streaming algorithms. More importantly, review work discussed in Section~\ref{sec:surveys} use the batch setting for evaluation, even though many algorithms under consideration, such as $k$-Means and CNNs, two basic and well studied ML algorithms that perform well in some TAD scenarios~\citep{liu2024elephant}, are algorithmically capable of addressing the streaming setting. As the evaluation strategy appropriate for streaming is different from the strategies employed in all the papers discussed in \ref{sec:surveys}, none of these actually provide guidance for practitioners. The two works that explicitly discuss this setting deploy rather niche algorithms, and don't evaluate on the wide array of current benchmark datasets, and don't consider common baselines.

To be more explicit, the evaluation in the streaming scenario for an algorithm $\mathcal{A}$ (producing a binary sequence of labels $\mathbf{a}$) on a time series $\mathbf{x}$ with ground truth labels $\mathbf{l}$ and a loss $\ell$, as discussed in \citet{boniol2021sand, ren2019time}\footnote{\citet{boniol2021sand} use minibatches, not individual points, and \citet{ren2019time} use an adjustment with a pre-specified delay tolerance.} can be formulated as
\begin{equation}
    R_o(\mathcal{A}, \mathbf{x}, \mathbf{l}) = \sum_t \ell\left(l_t, \mathcal{A}(\xtt)_t \right),
\end{equation}
i.e. computing the predicted anomaly label for each point $t$ using the algorithm on all data up to the point $t$.
This is in contrast to the batch evaluation used in all of the benchmarks described in Section~\ref{sec:surveys}, which use batch evaluation:
\begin{equation}
    R_b(\mathcal{A}, \mathbf{x}, \mathbf{l}) = \sum_t \ell\left(l_t, \mathcal{A}(\mathbf{x})_t \right),
\end{equation}
which allows fitting the algorithm once on the whole data set, and producing one set of predictions.

\subsubsection{Detection Delay}
While there has been much debate about metrics for time-series anomaly detection, with \citet{paparrizos2022volume} providing a comprehensive review as well as novel metrics, many of the recent proposals, such as Range-based Precision and Recall~\citep{tatbul2018precision} and VUS~\citep{paparrizos2022volume} take detection delay and detection window as a property to be invariant to.
For the streaming setting, in particular in time critical applications, the detection delay is actually an important aspect of algorithm performance, as described in \citet{lavin2015evaluating}. This differentiation has been noted in \citet{sorbo2024navigating}, but has been mostly ignored by other recent benchmarks.

\subsection{Side Information}\label{sec:side_information}
The pattern described in Section~\ref{sec:fridge}, where there is a population of time series of interest, and additional side-information attached to each time series is extremely common in practice.
There are two cases we want to distinguish: time-independent and time-varying side information.

\subsubsection{Time-independent Side Information: Subpopulation Mining}\label{sec:cohort_mining}
Time-independent side-information is often high-dimensional or categorical; for example, monitoring app-usage, each individual user might be associated with a country, a network provider, an operating system and a manufacturer. It could happen that suddenly traffic is altered because a patch rolled out by a phone manufacturer interacts badly with the routing policies of a national network provider\footnote{This is a sketch of a real example, however similar incidents are common in any large-scale IT infrastructure and often have dedicated teams to investigate}. As a streaming movie provider, you might find that suddenly interest in a certain actor or director spikes or decreases based on location and demographic factors, maybe because a news outlet has featured a story. In the cooling scenario described in \ref{sec:fridge}, we might have the manufacturer, years since installation, and state and warehouse ID as time-independent characteristics.

Formally, given a family of $d$ time series of length $n$, $x_{i,t}, t=1,\dots,n, i=1,\dots, d$, each associated with some time-independent features $\theta_i$, we want to find a rule or classifier $f_t$, such that $x_{i, t}$ is anomalous iff $f_t(\theta_i) > 0$.
Note that we want $f_t$ to be simple as a function of $\theta$, often just a conjunction of univariate terms, i.e. ``time series i belongs to device A in region B``, while the rules $f_t$ could vary arbitrarily over time $t$.
A widely used approach in practice is to apply an anomaly detection algorithm $\mathcal{A}$ to obtain anomaly labels $\mathbf{a}_{i, t}$ and then run a generic rule-mining approach $\mathcal{F}$ to find $f_t = \mathcal{F}((a_i, \theta_i)_{i=1,\dots, n})$ for each $t$ independently.
However, treating the anomaly detection problem jointly over instances, taking sub-populations into account, could inform the difficult univariate unsupervised TAD problem.

Modeling each individual time series could lead to problems, though; very fine grained modeling of a process can lead to very sparse signals that might need to be aggregated, see Section~\ref{sec:pointprocesses}, and to potentially very high-dimensional rule-mining spaces.

\subsubsection{Time-dependent Side Information: Conditional Anomalies}\label{sec:conditional_tad}
The other form of side information is including time-varying anomaly detection, which is closely related, but separate, from multivariate anomaly detection. In any application, no time-series stands alone, and sales, traffic or electricity usage depend on external factors like day of the week, holidays, temperature, interest rates etc.
Given multiple time-series $\mathbf{x}, \mathbf{y}, \mathbf{z}$, the question most commonly asked in the TAD literature is:
\begin{equation}\label{eq:multivariate_joint}
    \min_t p(x_t, y_t, z_t | \xtt, \ytt, \ztt)
\end{equation}
that is to find anomalies in the joint distribution of all variables at time $t$, given previous values, i.e. multivariate anomaly detection.
%
%
However the most practically relevant question is usually:
\begin{equation}\label{eq:multivariate_dependent}
    \min_t p(x_t| \xtt, \mathbf{y}_{<t+1}, \mathbf{z}_{<t+1})    
\end{equation}
i.e. find anomalies in $x_t$ given the values of the other series at that time and all previous timesteps. We call these \emph{conditional} anomalies.

To illustrate the difference between the notions, consider the classic mock example of the interaction of temperature $x_t$ and ice cream sales $y_t$. Let's assume they are highly related, and that heat-waves (which extremely high ice-cream sales) are rare.
Under these assumptions, a heat-wave would be an anomaly w.r.t \eqref{eq:multivariate_joint} (as the likelyhood of a heatwave is low), and not an anomaly w.r.t.
\eqref{eq:multivariate_dependent} 
as the relationship between ice-cream sales and temperature is maintained.
Now consider an event in which a very high ice-cream sale happens for another reason (a circus came to town, but we don't have corresponding data).
This scenario is an anomaly according to both \eqref{eq:multivariate_joint} and \eqref{eq:multivariate_dependent}.
Both of these formulations could be relevant, depending on the practical setting, while the literature only discusses \eqref{eq:multivariate_joint}. From our experience, in most scenarios, the two quantities of interest are anomalies in $p(x_t)$ and $p(x_t| \xtt, \mathbf{y}_{<t+1}, \mathbf{z}_{<t+1})$, which detect univariate anomalies, and univariate anomalies that are not explained by other factors, respectively.
As another illustrative example, consider monitoring flu cases as $x_t$, which are highly correlated with temperature $y_t$. We might be interested in whether there are abnormally many flu cases, i.e. low $p(x_t| \xtt)$, or we might be interested whether there are more flu cases than explained by cold temperatures, i.e. low $p(x_t| \xtt, \mathbf{y}_{<t+1})$. But we are likely not interested in a sudden heatwave or arctic vortexes, i.e. low $p(y_t| \ytt)$.
While there are many ways to address these problem of finding anomalies using \eqref{eq:multivariate_dependent}, say using residuals of a regression model or a quantile regression model, we advocate for a broader study of this area.
The challenge here is both in the availablility of datasets as well as methods; however, it might be possible to reuse data from existing multivariate benchmarks and relabel them to reflect this different setting.
We have not found a reference to the definition in Equation~\eqref{eq:multivariate_dependent} in the literature, which is somewhat surprising, as this seems to be the most commonly encountered scenario in practice, together with the time-independent information considered in Section~\ref{sec:cohort_mining}.

\subsection{Explainabilty and Root-Cause Analysis}
There are two layers of explainability that are commonly required in alerting. The first is ``Why was an alert triggered?'' and the second is ``What is the cause of the alert?''. These two questions require different tools and different levels of analysis. The first question can be posed and possibly answered even in the simplest case of a single univariate time series; it is a question of interpreting the anomaly-score provided by the algorithm. Essentially, if the score can be explained to the user, the alert is explainable. A simple threshold, say ``two standard deviations from the mean'' is easy to explain and can easily be depicted in a chart. Using an explainable forecasting method as the basis of anomaly detection is also relatively straight-forward: using exponential smoothing or possibly a simple autoregressive model, we can show the value that was forecast, explain why this forecast was made, and show that the difference to the forecast crossed a threshold.
Some discord-based methods can also be explained: ''The closest window we saw so far is this, and the distance crosses a threshold.`` Unfortunately none of the works described in Section~\ref{sec:surveys} consider this important axis of interpretability. Interpretability might not be absolute, and measuring the interpretability of methods would be a worthwhile survey; black-box forecasting methods like deep learning methods might or might not be considered explainable, depending on whether the forecast makes intuitive sense to users. A $k$-Means based approach might possibly be explainable, depending on the scenario. Anomaly scores directly produced by methods such as IForest~\cite{liu2008isolation} or Spectral Residual~\citep{ren2019time} offer limited explainability.

The second level of interpretability is usually related to external factors, ideally with a causal relationship, and can not usually be answered based on a single univariate series. Both time-independent and time-dependent side information, as described in Section~\ref{sec:side_information} could be considered to answer such a ``why'' question. Given the broad scope of this question, we will not discuss it in detail here, but refer the interested reader to \cite{yang2022causal} for one possible formulation of the problem. Solving this why question in a general sense could be seen as the holy grail of alerting; however, it is likely hard to solve robustly and in general, and we expect many of the more fundamental problems described in this paper are more likely to yield direct applications.

\section{Contextualizing and Human in the Loop}\label{sec:contextual}
As described in Section~\ref{sec:fridge}, there might be different kinds of anomalies present in the data, corresponding to different causes or physical effects, that might be relevant to different usage scenarios. More generally, there could be an arbitrary number of ways that a series could deviate from their usual pattern, only a sub-set of which might be relevant. In many applications, suppressing spurious alerts is critical for user trust and satisfaction, and efficient automation. Imagine a system in which false alerts about irrelevant behavior are sent to the fridge operators, who are sent to rescue food that is perfectly safe. Any alerting feature that has high false-positive rate is likely to be ignored soon. This could be addressed from a machine learning perspective by using a supervised system of online or active learning on top of unsupervised anomaly detection to decouple these two tasks. However, given the difficulty of unsupervised AD, it is likely that an integration of the two learning systems can yield substantial benefits.
\citet{liu2024elephant} and \citet{sorbo2024navigating} argue that inconsistent annotation in existing benchmarks creates a problem for evaluating algorithms; and while several benchmarks might have human error, another explanation is that not only that side-information required to determine anomalies is not included in the benchmark as time-varying features, but that side-information is present as context specific properties of the signal, only present in the operators mental model. While other mechanisms might be possible, the most obvious way to solve this problem is by using human annotation in the TAD process.

This problem could be cast as a classifying each point in a time series; however, given the extremely imbalanced labeling and sparse annotations, we argue that modeling this class of problems as its own task is more appropriate. Unfortunately, there are few datasets available for this task. The most highly curated benchmark, introduced in \citet{wu2021current}, only contains a single anomaly per time series, preventing any supervised or weakly supervised learning.
While modeling the problem of including human annotations using a batch setting with training and test set is possible, and was investigated in \citet{liu2015opprentice} we think that practical applications most closely resemble the online learning or active learning settings.
One of the differences compared to standard online or active learning is that in the TAD steaming setting, the feedback is usually one-sided, i.e. humans would annotate false positives, but not provide labels for false negatives, leading to partial feedback resembling a Bandit setting. Given that alerting usually includes a human recipient that will investigate each alert, we focus on the online setting.
The online setting can be formalized as finding an algorithm $\mathcal{A}$ that, given a sequence $\mathbf{x}$ and a set of labels $L_{<t}$ produces a binary outcome $a_t = \mathcal{A}(\xtt, L_{<t}) \ \in \{0, 1\}$ predicting whether $x_t$ is anomalous or not\footnote{While thresholding continuous anomaly scores is an  important topic, see \ref{sec:thresholdsetting}, we assume binary labels here for simplicity.}, with 
\[L_{<t}=\big \{(i, l_i)\; \big |\; \mathcal{A}(\mathbf{x}_{<i+1}, L_{<i}) = 1, i \in \{1, \dots, t-1\big \}.\]
In other words, the algorithm $\mathcal{A}$ has as input the time series $x$ up to time $t$, together with the true labels $l_t$ for all time points for which $\mathcal{A}(x_t) = 1$, i.e. that were flagged as anomalous before.
Then the regret to be minimized is
\begin{equation}
    R_\text{hil}(\mathcal{A}, \mathbf{x}, \mathbf{l}) = \sum_{i=1}^n \lambda_\text{fn} 1_{a_t = 0 \land l_t = 1}  + \lambda_\text{fp} 1_{a_t = 1 \land l_t = 0} 
\end{equation}
where $\lambda_\text{fn}$ and $\lambda_\text{fp}$ are application-specific costs for false negatives and false positives.
A related problem of censored feedback in multi-armed bandits was studied in \citet{abernethy2016threshold}, but we are not aware of a formulation with deterministic censorship based on the selected arm.
Alternatively, this could be seen as a reinforcement learning setting in an augmented state space in which the state is the pair $(\mathbf{x}_{<t+1}, L_{<t})$.

While the above seems the most straight-forward formulation of univariate anomaly detection as it is encountered in practice, we are not aware of prior work using this formulation, and invite the research community to consider this approach.



\begin{figure}[ht]
    \centering
    \begin{subfigure}[b]{\columnwidth}
        \centering
        \includegraphics[width=\textwidth]{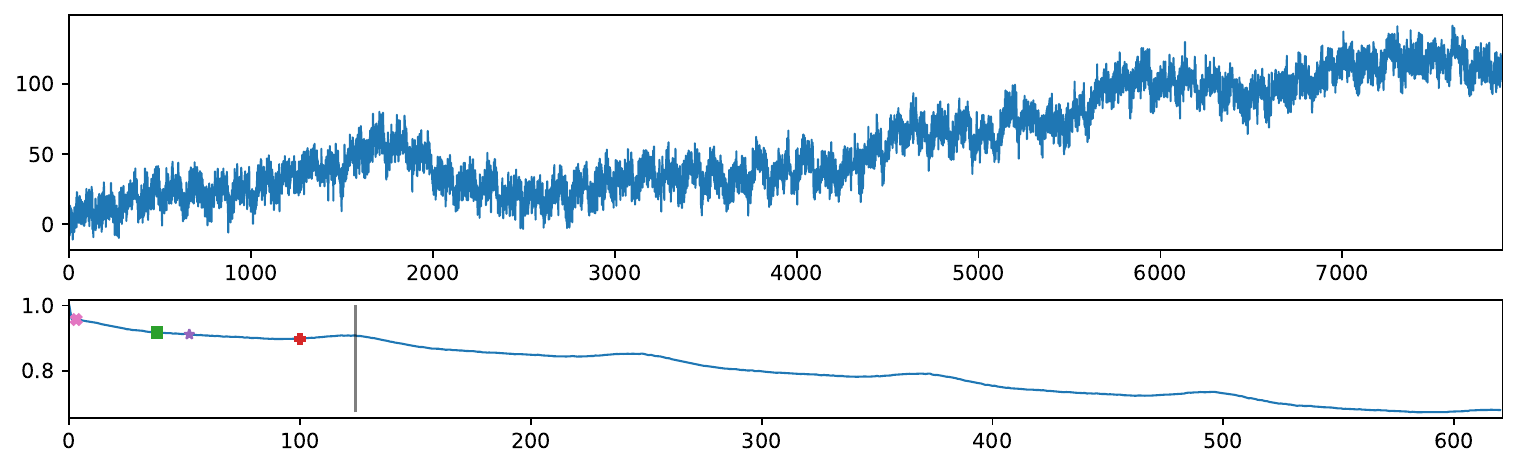}
    \end{subfigure}

    \begin{subfigure}[b]{\columnwidth}
        \centering
        \includegraphics[width=.8\textwidth]{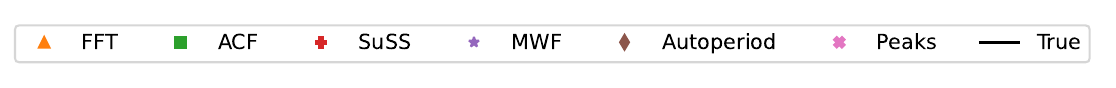}
    \end{subfigure}
    \caption{A synthetic time series (top),  on which the periodicity can easily be read from the autocorrelation function (bottom), but on which current methods fail, see markers on the autocorrelation function.}
    \label{fig:period_failure}
\end{figure}

\section{Preprocessing and Normalization}\label{sec:signalprocessing}
\subsection{Periodicity Detection and Window Size}
Algorithms for periodicity detection, in particular for multiple periodicities, still fail on cases that can be solved by manual inspection, leaving much room for improvement, see Figure~\ref{fig:period_failure} for an example comparing the methods investigated in \citet{ermshaus2023window} on a simple synthetic benchmark; see Appendix~\ref{app:window_size} for details. On the other hand, it is not entirely clear whether the periodicity of a signal is the right choice for the window size of TAD algorithms; while this choice is intuitive, benchmarks have shown fixed values of $4$ for PCA window size~\citep{schmidl2022anomaly} and $125$~(used for many sequences by default in \citet{liu2024elephant, paparrizos2022tsb}) to be very effective.

\begin{figure*}[ht]
    \centering
    \begin{subfigure}[b]{\textwidth}
        \centering
        \includegraphics[width=\textwidth]{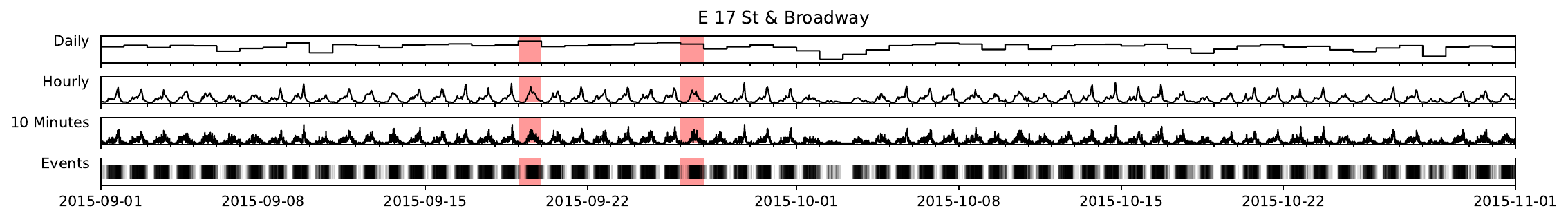}
    \end{subfigure}

    \begin{subfigure}[b]{\textwidth}
        \centering
        \includegraphics[width=\textwidth]{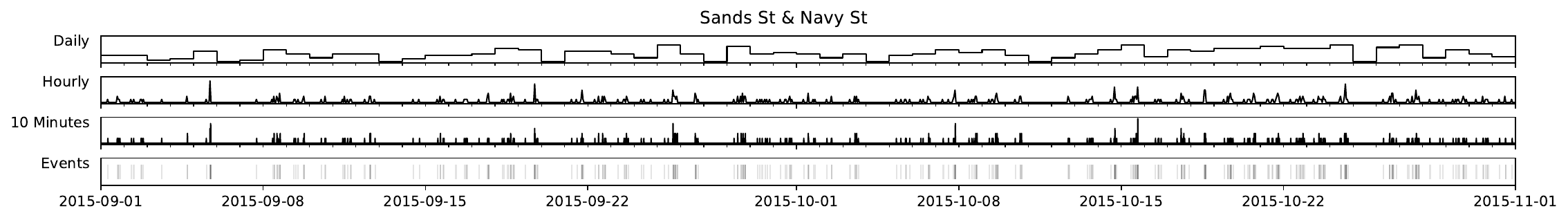}
    \end{subfigure}
    \caption{New York City citibike rental data from fall of 2015 for two different stations. The data is present as a point process and needs to be resampled to a time series for anomaly detection. Different downsampling rates might be appropriate for different stations, and it's non-obivous how to determine them automatically.}
    \label{fig:bikesdata}
\end{figure*}

\subsection{Point Processes and Resampling}\label{sec:pointprocesses}
In many IOT and other sensor scenarios, data does not form a time series, but instead is more naturally represented as a point process, for example measuring a sale or a person walking through a turnstile, or a visit to a website. Essentially all algorithms studied in the literature require a regular time series, however, and so resampling is required. Unfortunately, all current AD benchmarks contain regular time series, assuming this preprocessing has already been done. However, it's not obvious how to resample a signal in a way that does not smooth over existing anomalies or introduce new anomalies through sampling artifacts. An example is shown in Figure~\ref{fig:bikesdata} which shows citibike rental from fall 2015 for two different stations\footnote{Available at \url{https://citibikenyc.com/system-data}}. The data is stored as eventstream; however the number of events is very different for very busy stations (top) and slow stations (bottom). The figure shows three different resample rates: days, hours, and 10 minutes. For the busy station, all three gicve relatively clear signals, while for the slow station, only the daily resampling seems appropriate. However, using the daily rate for both stations would have a clear downside; on the top, an outlier is marked in red, showing two days with abnormal traffic patterns. Throughout the year, this station is most busy on workday evenings, though on these two days, the station is busy on a weekend morning. This is a strong anomaly on the hourly data, but can no longer be detected if aggregated to the daily level. We are not aware of the TAD literature discussing the issue of sampling rate determination at all, despite it being prominent in applications.
This issue is partially tied to the population issue discussed in Section~\ref{sec:cohort_mining}; looking at the sales for an individual product in a particular store might create an extremely sparse point process that could be hard to resample, while aggregating sales either across stores or products, or even both, would result in a much denser signal, that might be easier to process. Even among different products, sales might differ by orders of magnitude, requiring different resampling strategies depending on the popularity of a product, or aggregating some products into sub-categories to allow effective modeling. Many real-world processes follow long-tail distributions, such as the popularity of products, songs, movies and websites, which potentially requires different strategies for members of the same population.
Another hurdle that might seem trivial but can lead to  complications in practice is that the aggregation required on a point process is not obvious from the signal. A temperature measurement reading of 18$^\circ$ C at 10:10h and 19$^\circ$ C at 10:20h resampled to hourly data should lead to 18.5$^\circ$ C at 10:00h, but a \$18 sale at 10:10h and a \$19 sale at 10:20h means a sale of \$37 if resampled to the hour.

\subsection{Threshold Setting}\label{sec:thresholdsetting}
In an alerting setting, the choice of threshold for converting an anomaly score into an anomaly prediction is critical, since whether or not to send an alert is a binary decision. In a streaming setting, defining a single threshold might be difficult, and full information of the whole signal is not available. In contrast to these limitations, evaluation in benchmarks often uses an ideal threshold that not only requires knowledge of the full time series, but also of the ground truth~\citep{sorbo2024navigating}. In the streaming setting, \citet{hundman2018detecting} and \citet{ren2019time} both define ad-hoc thresholding strategies, but unfurtunately neither has been further investigate in the literature.


\section{Alternative Views}\label{sec:alternativeviews}
In this section we inspect possible counter-arguments and alternative points of view on our assumptions.

\subsection{Streaming Algorithms are not Necessary}
There is a potential caveat to the importance of streaming for TAD. While alerting is usually done on a live data stream, the frequency of the data is relatively low, often daily or hourly, and rarely less than several minutes.
Given these time frames it can be feasible to refit models from scratch at each time step. Depending on the scale of the application, this might or might not lead to infeasible computational costs (say refitting models for 1 million time series with three years of hourly data vs refitting a single model with several weeks of hourly data).  However, the coarser the resolution, the more important it is to have small alerting delays; if one algorithm alerts on the day of an anomaly, and another requires three more days of data to fill a window, this difference can be extremely consequential for applications.

\subsection{Supervision is not Required}\label{sec:no_supervision}
While we propose the viewpoint that a purely unsupervised approach to anomaly detection, in particular in the highly popular alerting setting, is sub-optimal, that does not imply it is entirely impractical. \citet{sylligardos2023choose} show that it is possible to select a TAD algorithm for a particular series without any additional annotation; however, the results are not outperforming a naive ensemble of all algorithms in an out-of-distribution setting.
Several surveys have found that different TAD algorithms are best suited for different datasets~\citep{liu2024elephant, schmidl2022anomaly}\todo{maybe plot, maybe appendix?}.
The approach of \citet{sylligardos2023choose} can only be effective if the information about which algorithm is best suited to a time series is intrinsic to the signal encoded in the time series itself. Only then can a priori decisions about the algorithm choice be made. 
Our second tenet contradicts this assumption.
Whether our requirement holds true on existing academic benchmarks is yet to be determined; the difference between in-distribution and out-of-distribution performance observed in \citet{sylligardos2023choose} might be a clue that the nature of the annotation within a dataset, and not only the signal characteristics, are required to choose an algorithm. On the other hand, better algorithms or better features could yield better algorithm selection, which would lead to the conclusion that most of the information required to determine anomalies is indeed contained in the signal itself.
We are convinced that determining the application-relevant anomalies just from the signal is impossible; however, assuming a reasonable superset can be identified, this might be enough for some applications. If an anomaly occurs every 100,000 observations, but only half of those are relevant for a specific application, this would still reduce manual labor by orders of magnitude, even if precision could not exceed a certain level (0.5 in this example).
Rigorously proving our tenet might be difficult; however, given appropriate datasets, it seems worthwhile to investigate how much we can gain by incorporating human knowledge.

\subsection{Anomaly Semantics Could be Made Explicit}
An alternative approach to manual annotation with a human in the loop as described in Section~\ref{sec:contextual} would be to include expert knowledge without annotating the data, for example via hyper-parameter or algorithm choice. 
\citet{schmidl2022anomaly} show that properties of the anomaly type determine the accuracy of different algorithms, at least to a certain degree, and it is possible that with a better understanding, we can make qualitative recommendations to users based on their understanding of the data. Executing this strategy, however, requires communicating the relevant categories of anomalies to the user to make the correct determination; whether it is possible is an open question.

\subsection{Metrics and Annotations are to Blame}
Another possible explanation of the diversity of results, and the difficulties in ranking algorithms and providing consistent results is given by \citet{sorbo2024navigating}; they conclude ``many metrics are not appropriate for certain kinds of labelling strategies'', which implies that labeling strategy and metric need to be matched. Given the many different labeling strategies employed for different datasets~\citep{liu2024elephant}, and the patterns rewarded or penalized by different metrics, this rings true. However, this stance distributes the burden of providing accurate results between annotations, metrics and algorithms, making it extremely difficult for a practitioner to know what combination is most appropriate for their specific application. \citet{liu2024elephant} made an attempt in unifying annotations; however, our own preliminary experiments have shown rankings to still be very metric dependent.





\section{Conclusions}
This paper discussed two core tenets, \alerting and \applicationspecific, that we belief are the foundation of applying time series anomaly detection in practice. These lead us to a wide variety of problems that appear in practice that we believe are understudied in the ML community. While this work contains a long list of challenges, these can be summarized in three different categories: defining problems, solving problems and integrating solutions. While we attempted to define some of the problems, there is much work to be done to crisply define anomalies, RCA, or the different TAD settings we described. Based on these, there are clear theoretical and practical challenges to address these highly practically relevant but understudied settings. Finally, many aspects of the problem have been treated separately or ad-hoc, such as human feedback, signal processing and explanations, even though they are deeply interconnected, and more integrated approaches could improve the overall performance of the systems.
We hope that this paper will motivate others to investigate these problem settings.

\bibliographystyle{ACM-Reference-Format}
\bibliography{open_challenges_tad}


\begin{thebibliography}{40}


\ifx \showCODEN    \undefined \def \showCODEN     #1{\unskip}     \fi
\ifx \showISBNx    \undefined \def \showISBNx     #1{\unskip}     \fi
\ifx \showISBNxiii \undefined \def \showISBNxiii  #1{\unskip}     \fi
\ifx \showISSN     \undefined \def \showISSN      #1{\unskip}     \fi
\ifx \showLCCN     \undefined \def \showLCCN      #1{\unskip}     \fi
\ifx \shownote     \undefined \def \shownote      #1{#1}          \fi
\ifx \showarticletitle \undefined \def \showarticletitle #1{#1}   \fi
\ifx \showURL      \undefined \def \showURL       {\relax}        \fi
\providecommand\bibfield[2]{#2}
\providecommand\bibinfo[2]{#2}
\providecommand\natexlab[1]{#1}
\providecommand\showeprint[2][]{arXiv:#2}

\bibitem[Abernethy et~al\mbox{.}(2016)]%
        {abernethy2016threshold}
\bibfield{author}{\bibinfo{person}{Jacob~D Abernethy}, \bibinfo{person}{Kareem Amin}, {and} \bibinfo{person}{Ruihao Zhu}.} \bibinfo{year}{2016}\natexlab{}.
\newblock \showarticletitle{Threshold bandits, with and without censored feedback}.
\newblock \bibinfo{journal}{\emph{Advances In Neural Information Processing Systems}}  \bibinfo{volume}{29} (\bibinfo{year}{2016}).
\newblock


\bibitem[Boniol et~al\mbox{.}(2021)]%
        {boniol2021sand}
\bibfield{author}{\bibinfo{person}{Paul Boniol}, \bibinfo{person}{John Paparrizos}, \bibinfo{person}{Themis Palpanas}, {and} \bibinfo{person}{Michael~J Franklin}.} \bibinfo{year}{2021}\natexlab{}.
\newblock \showarticletitle{SAND: streaming subsequence anomaly detection}.
\newblock \bibinfo{journal}{\emph{Proceedings of the VLDB Endowment}} \bibinfo{volume}{14}, \bibinfo{number}{10} (\bibinfo{year}{2021}), \bibinfo{pages}{1717--1729}.
\newblock


\bibitem[Deng et~al\mbox{.}(2024)]%
        {deng2024reliable}
\bibfield{author}{\bibinfo{person}{Ziquan Deng}, \bibinfo{person}{Xiwei Xuan}, \bibinfo{person}{Kwan-Liu Ma}, {and} \bibinfo{person}{Zhaodan Kong}.} \bibinfo{year}{2024}\natexlab{}.
\newblock \showarticletitle{A Reliable Framework for Human-in-the-Loop Anomaly Detection in Time Series}.
\newblock \bibinfo{journal}{\emph{arXiv preprint arXiv:2405.03234}} (\bibinfo{year}{2024}).
\newblock


\bibitem[Elfeky et~al\mbox{.}(2005)]%
        {elfeky2005periodicity}
\bibfield{author}{\bibinfo{person}{Mohamed~G Elfeky}, \bibinfo{person}{Walid~G Aref}, {and} \bibinfo{person}{Ahmed~K Elmagarmid}.} \bibinfo{year}{2005}\natexlab{}.
\newblock \showarticletitle{Periodicity detection in time series databases}.
\newblock \bibinfo{journal}{\emph{IEEE Transactions on Knowledge and Data Engineering}} \bibinfo{volume}{17}, \bibinfo{number}{7} (\bibinfo{year}{2005}), \bibinfo{pages}{875--887}.
\newblock


\bibitem[Ermshaus et~al\mbox{.}(2023a)]%
        {ermshaus2023clasp}
\bibfield{author}{\bibinfo{person}{Arik Ermshaus}, \bibinfo{person}{Patrick Sch{\"a}fer}, {and} \bibinfo{person}{Ulf Leser}.} \bibinfo{year}{2023}\natexlab{a}.
\newblock \showarticletitle{ClaSP: parameter-free time series segmentation}.
\newblock \bibinfo{journal}{\emph{Data Mining and Knowledge Discovery}} \bibinfo{volume}{37}, \bibinfo{number}{3} (\bibinfo{year}{2023}), \bibinfo{pages}{1262--1300}.
\newblock


\bibitem[Ermshaus et~al\mbox{.}(2023b)]%
        {ermshaus2023window}
\bibfield{author}{\bibinfo{person}{Arik Ermshaus}, \bibinfo{person}{Patrick Sch{\"a}fer}, {and} \bibinfo{person}{Ulf Leser}.} \bibinfo{year}{2023}\natexlab{b}.
\newblock \showarticletitle{Window size selection in unsupervised time series analytics: A review and benchmark}. In \bibinfo{booktitle}{\emph{International Workshop on Advanced Analytics and Learning on Temporal Data}}. Springer, \bibinfo{pages}{83--101}.
\newblock


\bibitem[Gungor et~al\mbox{.}(2025)]%
        {gungor2025robust}
\bibfield{author}{\bibinfo{person}{Onat Gungor}, \bibinfo{person}{Amanda Rios}, \bibinfo{person}{Priyanka Mudgal}, \bibinfo{person}{Nilesh Ahuja}, {and} \bibinfo{person}{Tajana Rosing}.} \bibinfo{year}{2025}\natexlab{}.
\newblock \showarticletitle{A Robust Framework for Evaluation of Unsupervised Time-Series Anomaly Detection}. In \bibinfo{booktitle}{\emph{International Conference on Pattern Recognition}}. Springer, \bibinfo{pages}{48--64}.
\newblock


\bibitem[Hilal et~al\mbox{.}(2022)]%
        {hilal2022financial}
\bibfield{author}{\bibinfo{person}{Waleed Hilal}, \bibinfo{person}{S~Andrew Gadsden}, {and} \bibinfo{person}{John Yawney}.} \bibinfo{year}{2022}\natexlab{}.
\newblock \showarticletitle{Financial fraud: a review of anomaly detection techniques and recent advances}.
\newblock \bibinfo{journal}{\emph{Expert systems With applications}}  \bibinfo{volume}{193} (\bibinfo{year}{2022}), \bibinfo{pages}{116429}.
\newblock


\bibitem[Huang et~al\mbox{.}(2023)]%
        {huang2023labelled}
\bibfield{author}{\bibinfo{person}{Tao Huang}, \bibinfo{person}{Silas N{\o}stvik}, \bibinfo{person}{Peder Bacher}, \bibinfo{person}{Jonas~Kj{\ae}r Jensen}, \bibinfo{person}{Wiebke~Brix Markussen}, {and} \bibinfo{person}{Jan~Kloppenborg M{\o}ller}.} \bibinfo{year}{2023}\natexlab{}.
\newblock \showarticletitle{Labelled dataset for Ultra-Low Temperature Freezer to aid dynamic modelling \& fault detection and diagnostics}.
\newblock \bibinfo{journal}{\emph{Scientific Data}} \bibinfo{volume}{10}, \bibinfo{number}{1} (\bibinfo{year}{2023}), \bibinfo{pages}{888}.
\newblock


\bibitem[Hundman et~al\mbox{.}(2018)]%
        {hundman2018detecting}
\bibfield{author}{\bibinfo{person}{Kyle Hundman}, \bibinfo{person}{Valentino Constantinou}, \bibinfo{person}{Christopher Laporte}, \bibinfo{person}{Ian Colwell}, {and} \bibinfo{person}{Tom Soderstrom}.} \bibinfo{year}{2018}\natexlab{}.
\newblock \showarticletitle{Detecting spacecraft anomalies using lstms and nonparametric dynamic thresholding}. In \bibinfo{booktitle}{\emph{Proceedings of the 24th ACM SIGKDD international conference on knowledge discovery \& data mining}}. \bibinfo{pages}{387--395}.
\newblock


\bibitem[Imani and Keogh(2021)]%
        {imani2021multi}
\bibfield{author}{\bibinfo{person}{Shima Imani} {and} \bibinfo{person}{Eamonn Keogh}.} \bibinfo{year}{2021}\natexlab{}.
\newblock \showarticletitle{Multi-window-finder: domain agnostic window size for time series data}.
\newblock \bibinfo{journal}{\emph{Proceedings of the MileTS}}  \bibinfo{volume}{21} (\bibinfo{year}{2021}).
\newblock


\bibitem[James et~al\mbox{.}(2016)]%
        {james2016leveraging}
\bibfield{author}{\bibinfo{person}{Nicholas~A James}, \bibinfo{person}{Arun Kejariwal}, {and} \bibinfo{person}{David~S Matteson}.} \bibinfo{year}{2016}\natexlab{}.
\newblock \showarticletitle{Leveraging cloud data to mitigate user experience from ‘breaking bad’}. In \bibinfo{booktitle}{\emph{2016 IEEE International Conference on Big Data (Big Data)}}. IEEE, \bibinfo{pages}{3499--3508}.
\newblock


\bibitem[Kim et~al\mbox{.}(2022)]%
        {kim2022towards}
\bibfield{author}{\bibinfo{person}{Siwon Kim}, \bibinfo{person}{Kukjin Choi}, \bibinfo{person}{Hyun-Soo Choi}, \bibinfo{person}{Byunghan Lee}, {and} \bibinfo{person}{Sungroh Yoon}.} \bibinfo{year}{2022}\natexlab{}.
\newblock \showarticletitle{Towards a rigorous evaluation of time-series anomaly detection}. In \bibinfo{booktitle}{\emph{Proceedings of the AAAI Conference on Artificial Intelligence}}, Vol.~\bibinfo{volume}{36}. \bibinfo{pages}{7194--7201}.
\newblock


\bibitem[Lai et~al\mbox{.}(2021)]%
        {lai2021revisiting}
\bibfield{author}{\bibinfo{person}{Kwei-Herng Lai}, \bibinfo{person}{Daochen Zha}, \bibinfo{person}{Junjie Xu}, \bibinfo{person}{Yue Zhao}, \bibinfo{person}{Guanchu Wang}, {and} \bibinfo{person}{Xia Hu}.} \bibinfo{year}{2021}\natexlab{}.
\newblock \showarticletitle{Revisiting time series outlier detection: Definitions and benchmarks}. In \bibinfo{booktitle}{\emph{Thirty-fifth conference on neural information processing systems datasets and benchmarks track (round 1)}}.
\newblock


\bibitem[Laptev et~al\mbox{.}(2015)]%
        {laptev2015generic}
\bibfield{author}{\bibinfo{person}{Nikolay Laptev}, \bibinfo{person}{Saeed Amizadeh}, {and} \bibinfo{person}{Ian Flint}.} \bibinfo{year}{2015}\natexlab{}.
\newblock \showarticletitle{Generic and scalable framework for automated time-series anomaly detection}. In \bibinfo{booktitle}{\emph{Proceedings of the 21th ACM SIGKDD international conference on knowledge discovery and data mining}}. \bibinfo{pages}{1939--1947}.
\newblock


\bibitem[Lavin and Ahmad(2015)]%
        {lavin2015evaluating}
\bibfield{author}{\bibinfo{person}{Alexander Lavin} {and} \bibinfo{person}{Subutai Ahmad}.} \bibinfo{year}{2015}\natexlab{}.
\newblock \showarticletitle{Evaluating real-time anomaly detection algorithms--the Numenta anomaly benchmark}. In \bibinfo{booktitle}{\emph{2015 IEEE 14th international conference on machine learning and applications (ICMLA)}}. IEEE, \bibinfo{pages}{38--44}.
\newblock


\bibitem[Liu et~al\mbox{.}(2015)]%
        {liu2015opprentice}
\bibfield{author}{\bibinfo{person}{Dapeng Liu}, \bibinfo{person}{Youjian Zhao}, \bibinfo{person}{Haowen Xu}, \bibinfo{person}{Yongqian Sun}, \bibinfo{person}{Dan Pei}, \bibinfo{person}{Jiao Luo}, \bibinfo{person}{Xiaowei Jing}, {and} \bibinfo{person}{Mei Feng}.} \bibinfo{year}{2015}\natexlab{}.
\newblock \showarticletitle{Opprentice: Towards practical and automatic anomaly detection through machine learning}. In \bibinfo{booktitle}{\emph{Proceedings of the 2015 internet measurement conference}}. \bibinfo{pages}{211--224}.
\newblock


\bibitem[Liu et~al\mbox{.}(2008)]%
        {liu2008isolation}
\bibfield{author}{\bibinfo{person}{Fei~Tony Liu}, \bibinfo{person}{Kai~Ming Ting}, {and} \bibinfo{person}{Zhi-Hua Zhou}.} \bibinfo{year}{2008}\natexlab{}.
\newblock \showarticletitle{Isolation forest}. In \bibinfo{booktitle}{\emph{2008 eighth ieee international conference on data mining}}. IEEE, \bibinfo{pages}{413--422}.
\newblock


\bibitem[Liu et~al\mbox{.}(2024)]%
        {liu2024time}
\bibfield{author}{\bibinfo{person}{Qinghua Liu}, \bibinfo{person}{Paul Boniol}, \bibinfo{person}{Themis Palpanas}, {and} \bibinfo{person}{John Paparrizos}.} \bibinfo{year}{2024}\natexlab{}.
\newblock \showarticletitle{Time-Series Anomaly Detection: Overview and New Trends}.
\newblock \bibinfo{journal}{\emph{Proceedings of the VLDB Endowment (PVLDB)}} \bibinfo{volume}{17}, \bibinfo{number}{12} (\bibinfo{year}{2024}), \bibinfo{pages}{4229--4232}.
\newblock


\bibitem[Liu and Paparrizos(2024)]%
        {liu2024elephant}
\bibfield{author}{\bibinfo{person}{Qinghua Liu} {and} \bibinfo{person}{John Paparrizos}.} \bibinfo{year}{2024}\natexlab{}.
\newblock \showarticletitle{The Elephant in the Room: Towards A Reliable Time-Series Anomaly Detection Benchmark}. In \bibinfo{booktitle}{\emph{The Thirty-eight Conference on Neural Information Processing Systems Datasets and Benchmarks Track}}.
\newblock


\bibitem[Mejri et~al\mbox{.}(2024)]%
        {mejri2024unsupervised}
\bibfield{author}{\bibinfo{person}{Nesryne Mejri}, \bibinfo{person}{Laura Lopez-Fuentes}, \bibinfo{person}{Kankana Roy}, \bibinfo{person}{Pavel Chernakov}, \bibinfo{person}{Enjie Ghorbel}, {and} \bibinfo{person}{Djamila Aouada}.} \bibinfo{year}{2024}\natexlab{}.
\newblock \showarticletitle{Unsupervised anomaly detection in time-series: An extensive evaluation and analysis of state-of-the-art methods}.
\newblock \bibinfo{journal}{\emph{Expert Systems with Applications}} (\bibinfo{year}{2024}), \bibinfo{pages}{124922}.
\newblock


\bibitem[Oppenheim(1999)]%
        {oppenheim1999discrete}
\bibfield{author}{\bibinfo{person}{Alan~V Oppenheim}.} \bibinfo{year}{1999}\natexlab{}.
\newblock \bibinfo{booktitle}{\emph{Discrete-time signal processing}}.
\newblock \bibinfo{publisher}{Pearson Education India}.
\newblock


\bibitem[Paparrizos et~al\mbox{.}(2022a)]%
        {paparrizos2022volume}
\bibfield{author}{\bibinfo{person}{John Paparrizos}, \bibinfo{person}{Paul Boniol}, \bibinfo{person}{Themis Palpanas}, \bibinfo{person}{Ruey~S Tsay}, \bibinfo{person}{Aaron Elmore}, {and} \bibinfo{person}{Michael~J Franklin}.} \bibinfo{year}{2022}\natexlab{a}.
\newblock \showarticletitle{Volume under the surface: a new accuracy evaluation measure for time-series anomaly detection}.
\newblock \bibinfo{journal}{\emph{Proceedings of the VLDB Endowment}} \bibinfo{volume}{15}, \bibinfo{number}{11} (\bibinfo{year}{2022}), \bibinfo{pages}{2774--2787}.
\newblock


\bibitem[Paparrizos et~al\mbox{.}(2022b)]%
        {paparrizos2022tsb}
\bibfield{author}{\bibinfo{person}{John Paparrizos}, \bibinfo{person}{Yuhao Kang}, \bibinfo{person}{Paul Boniol}, \bibinfo{person}{Ruey~S Tsay}, \bibinfo{person}{Themis Palpanas}, {and} \bibinfo{person}{Michael~J Franklin}.} \bibinfo{year}{2022}\natexlab{b}.
\newblock \showarticletitle{TSB-UAD: an end-to-end benchmark suite for univariate time-series anomaly detection}.
\newblock \bibinfo{journal}{\emph{Proceedings of the VLDB Endowment}} \bibinfo{volume}{15}, \bibinfo{number}{8} (\bibinfo{year}{2022}), \bibinfo{pages}{1697--1711}.
\newblock


\bibitem[Ren et~al\mbox{.}(2019)]%
        {ren2019time}
\bibfield{author}{\bibinfo{person}{Hansheng Ren}, \bibinfo{person}{Bixiong Xu}, \bibinfo{person}{Yujing Wang}, \bibinfo{person}{Chao Yi}, \bibinfo{person}{Congrui Huang}, \bibinfo{person}{Xiaoyu Kou}, \bibinfo{person}{Tony Xing}, \bibinfo{person}{Mao Yang}, \bibinfo{person}{Jie Tong}, {and} \bibinfo{person}{Qi Zhang}.} \bibinfo{year}{2019}\natexlab{}.
\newblock \showarticletitle{Time-series anomaly detection service at microsoft}. In \bibinfo{booktitle}{\emph{Proceedings of the 25th ACM SIGKDD international conference on knowledge discovery \& data mining}}. \bibinfo{pages}{3009--3017}.
\newblock


\bibitem[Rousseeuw and Driessen(1999)]%
        {rousseeuw1999fast}
\bibfield{author}{\bibinfo{person}{Peter~J Rousseeuw} {and} \bibinfo{person}{Katrien~Van Driessen}.} \bibinfo{year}{1999}\natexlab{}.
\newblock \showarticletitle{A fast algorithm for the minimum covariance determinant estimator}.
\newblock \bibinfo{journal}{\emph{Technometrics}} \bibinfo{volume}{41}, \bibinfo{number}{3} (\bibinfo{year}{1999}), \bibinfo{pages}{212--223}.
\newblock


\bibitem[Sarfraz et~al\mbox{.}(2024)]%
        {sarfrazposition}
\bibfield{author}{\bibinfo{person}{M~Saquib Sarfraz}, \bibinfo{person}{Mei-Yen Chen}, \bibinfo{person}{Lukas Layer}, \bibinfo{person}{Kunyu Peng}, {and} \bibinfo{person}{Marios Koulakis}.} \bibinfo{year}{2024}\natexlab{}.
\newblock \showarticletitle{Position: Quo Vadis, Unsupervised Time Series Anomaly Detection?}. In \bibinfo{booktitle}{\emph{Forty-first International Conference on Machine Learning}}.
\newblock


\bibitem[Schmidl et~al\mbox{.}(2022)]%
        {schmidl2022anomaly}
\bibfield{author}{\bibinfo{person}{Sebastian Schmidl}, \bibinfo{person}{Phillip Wenig}, {and} \bibinfo{person}{Thorsten Papenbrock}.} \bibinfo{year}{2022}\natexlab{}.
\newblock \showarticletitle{Anomaly detection in time series: a comprehensive evaluation}.
\newblock \bibinfo{journal}{\emph{Proceedings of the VLDB Endowment}} \bibinfo{volume}{15}, \bibinfo{number}{9} (\bibinfo{year}{2022}), \bibinfo{pages}{1779--1797}.
\newblock


\bibitem[Soldani and Brogi(2022)]%
        {soldani2022anomaly}
\bibfield{author}{\bibinfo{person}{Jacopo Soldani} {and} \bibinfo{person}{Antonio Brogi}.} \bibinfo{year}{2022}\natexlab{}.
\newblock \showarticletitle{Anomaly detection and failure root cause analysis in (micro) service-based cloud applications: A survey}.
\newblock \bibinfo{journal}{\emph{ACM Computing Surveys (CSUR)}} \bibinfo{volume}{55}, \bibinfo{number}{3} (\bibinfo{year}{2022}), \bibinfo{pages}{1--39}.
\newblock


\bibitem[S{\o}rb{\o} and Ruocco(2024)]%
        {sorbo2024navigating}
\bibfield{author}{\bibinfo{person}{Sondre S{\o}rb{\o}} {and} \bibinfo{person}{Massimiliano Ruocco}.} \bibinfo{year}{2024}\natexlab{}.
\newblock \showarticletitle{Navigating the metric maze: A taxonomy of evaluation metrics for anomaly detection in time series}.
\newblock \bibinfo{journal}{\emph{Data Mining and Knowledge Discovery}} \bibinfo{volume}{38}, \bibinfo{number}{3} (\bibinfo{year}{2024}), \bibinfo{pages}{1027--1068}.
\newblock


\bibitem[Sylligardos et~al\mbox{.}(2023)]%
        {sylligardos2023choose}
\bibfield{author}{\bibinfo{person}{Emmanouil Sylligardos}, \bibinfo{person}{Paul Boniol}, \bibinfo{person}{John Paparrizos}, \bibinfo{person}{Panos Trahanias}, {and} \bibinfo{person}{Themis Palpanas}.} \bibinfo{year}{2023}\natexlab{}.
\newblock \showarticletitle{Choose wisely: An extensive evaluation of model selection for anomaly detection in time series}.
\newblock \bibinfo{journal}{\emph{Proceedings of the VLDB Endowment}} \bibinfo{volume}{16}, \bibinfo{number}{11} (\bibinfo{year}{2023}), \bibinfo{pages}{3418--3432}.
\newblock


\bibitem[Tatbul et~al\mbox{.}(2018)]%
        {tatbul2018precision}
\bibfield{author}{\bibinfo{person}{Nesime Tatbul}, \bibinfo{person}{Tae~Jun Lee}, \bibinfo{person}{Stan Zdonik}, \bibinfo{person}{Mejbah Alam}, {and} \bibinfo{person}{Justin Gottschlich}.} \bibinfo{year}{2018}\natexlab{}.
\newblock \showarticletitle{Precision and Recall for Time Series}.
\newblock   \bibinfo{volume}{31} (\bibinfo{year}{2018}).
\newblock
\urldef\tempurl%
\url{https://proceedings.neurips.cc/paper_files/paper/2018/file/8f468c873a32bb0619eaeb2050ba45d1-Paper.pdf}
\showURL{%
\tempurl}


\bibitem[Ten et~al\mbox{.}(2011)]%
        {ten2011anomaly}
\bibfield{author}{\bibinfo{person}{Chee-Wooi Ten}, \bibinfo{person}{Junho Hong}, {and} \bibinfo{person}{Chen-Ching Liu}.} \bibinfo{year}{2011}\natexlab{}.
\newblock \showarticletitle{Anomaly detection for cybersecurity of the substations}.
\newblock \bibinfo{journal}{\emph{IEEE Transactions on Smart Grid}} \bibinfo{volume}{2}, \bibinfo{number}{4} (\bibinfo{year}{2011}), \bibinfo{pages}{865--873}.
\newblock


\bibitem[Vallis et~al\mbox{.}(2014)]%
        {vallis2014novel}
\bibfield{author}{\bibinfo{person}{Owen Vallis}, \bibinfo{person}{Jordan Hochenbaum}, {and} \bibinfo{person}{Arun Kejariwal}.} \bibinfo{year}{2014}\natexlab{}.
\newblock \showarticletitle{A novel technique for $\{$Long-Term$\}$ anomaly detection in the cloud}. In \bibinfo{booktitle}{\emph{6th USENIX workshop on hot topics in cloud computing (HotCloud 14)}}.
\newblock


\bibitem[Vlachos et~al\mbox{.}(2005)]%
        {vlachos2005periodicity}
\bibfield{author}{\bibinfo{person}{Michail Vlachos}, \bibinfo{person}{Philip Yu}, {and} \bibinfo{person}{Vittorio Castelli}.} \bibinfo{year}{2005}\natexlab{}.
\newblock \showarticletitle{On periodicity detection and structural periodic similarity}. In \bibinfo{booktitle}{\emph{Proceedings of the 2005 SIAM international conference on data mining}}. SIAM, \bibinfo{pages}{449--460}.
\newblock


\bibitem[Wen et~al\mbox{.}(2021)]%
        {wen2021robustperiod}
\bibfield{author}{\bibinfo{person}{Qingsong Wen}, \bibinfo{person}{Kai He}, \bibinfo{person}{Liang Sun}, \bibinfo{person}{Yingying Zhang}, \bibinfo{person}{Min Ke}, {and} \bibinfo{person}{Huan Xu}.} \bibinfo{year}{2021}\natexlab{}.
\newblock \showarticletitle{RobustPeriod: Robust time-frequency mining for multiple periodicity detection}. In \bibinfo{booktitle}{\emph{Proceedings of the 2021 international conference on management of data}}. \bibinfo{pages}{2328--2337}.
\newblock


\bibitem[Wu and Keogh(2021)]%
        {wu2021current}
\bibfield{author}{\bibinfo{person}{Renjie Wu} {and} \bibinfo{person}{Eamonn~J Keogh}.} \bibinfo{year}{2021}\natexlab{}.
\newblock \showarticletitle{Current time series anomaly detection benchmarks are flawed and are creating the illusion of progress}.
\newblock \bibinfo{journal}{\emph{IEEE transactions on knowledge and data engineering}} \bibinfo{volume}{35}, \bibinfo{number}{3} (\bibinfo{year}{2021}), \bibinfo{pages}{2421--2429}.
\newblock


\bibitem[Yang et~al\mbox{.}(2022)]%
        {yang2022causal}
\bibfield{author}{\bibinfo{person}{Wenzhuo Yang}, \bibinfo{person}{Kun Zhang}, {and} \bibinfo{person}{Steven~CH Hoi}.} \bibinfo{year}{2022}\natexlab{}.
\newblock \showarticletitle{A causal approach to detecting multivariate time-series anomalies and root causes}.
\newblock \bibinfo{journal}{\emph{arXiv preprint arXiv:2206.15033}} (\bibinfo{year}{2022}).
\newblock


\bibitem[Yu et~al\mbox{.}(2016)]%
        {yu2016advances}
\bibfield{author}{\bibinfo{person}{Zhun~Jerry Yu}, \bibinfo{person}{Fariborz Haghighat}, {and} \bibinfo{person}{Benjamin~CM Fung}.} \bibinfo{year}{2016}\natexlab{}.
\newblock \showarticletitle{Advances and challenges in building engineering and data mining applications for energy-efficient communities}.
\newblock \bibinfo{journal}{\emph{Sustainable Cities and Society}}  \bibinfo{volume}{25} (\bibinfo{year}{2016}), \bibinfo{pages}{33--38}.
\newblock


\bibitem[Zhang et~al\mbox{.}(2021)]%
        {zhang2021time}
\bibfield{author}{\bibinfo{person}{Jiuqi~Elise Zhang}, \bibinfo{person}{Di Wu}, {and} \bibinfo{person}{Benoit Boulet}.} \bibinfo{year}{2021}\natexlab{}.
\newblock \showarticletitle{Time series anomaly detection for smart grids: A survey}. In \bibinfo{booktitle}{\emph{2021 IEEE electrical power and energy conference (EPEC)}}. IEEE, \bibinfo{pages}{125--130}.
\newblock


\end{thebibliography}

\newpage

\appendix

\begin{figure}[ht]
    \centering
    \begin{subfigure}[b]{\columnwidth}
         \centering
         \includegraphics[width=\textwidth]{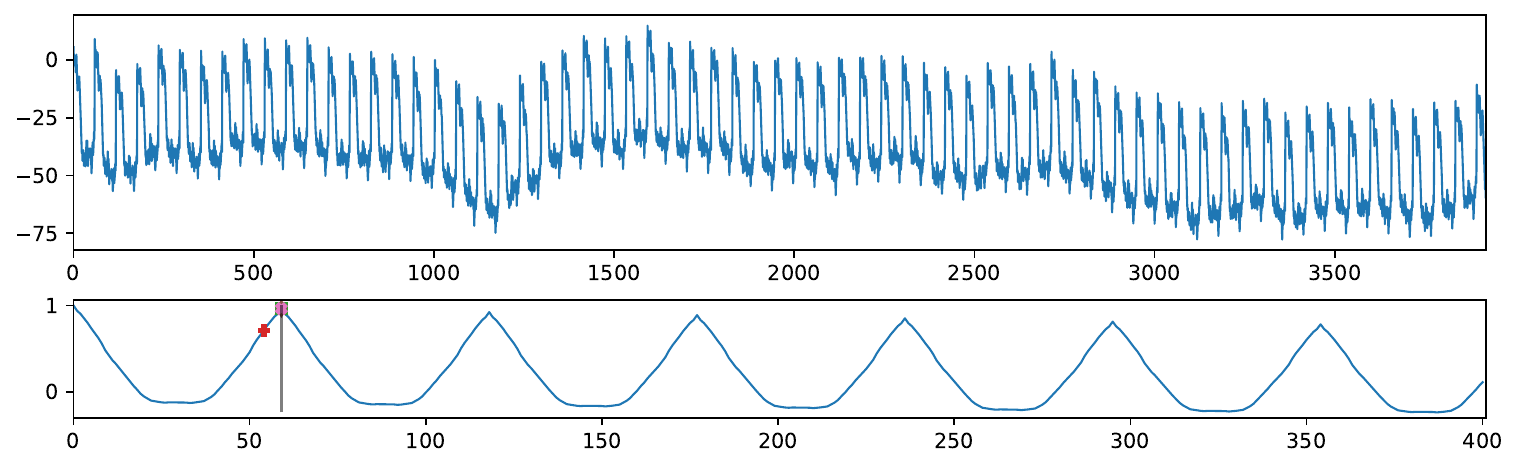}

    \end{subfigure}
    \begin{subfigure}[b]{\columnwidth}
         \centering
         \includegraphics[width=\textwidth]{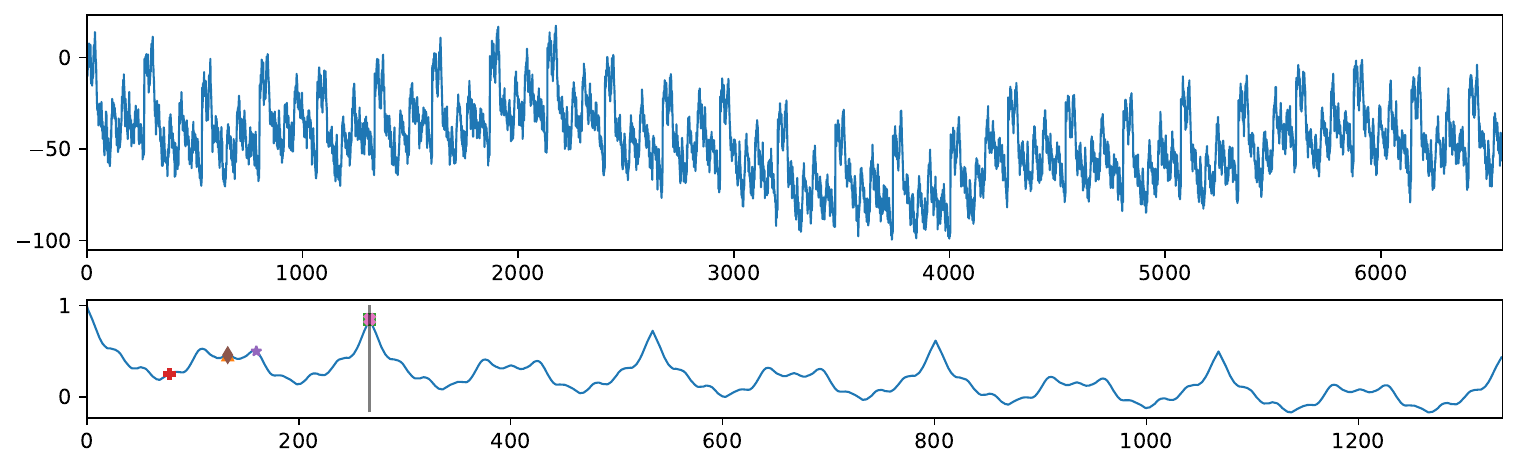}
    \end{subfigure}
    \begin{subfigure}[b]{\columnwidth}
         \centering
         \includegraphics[width=\textwidth]{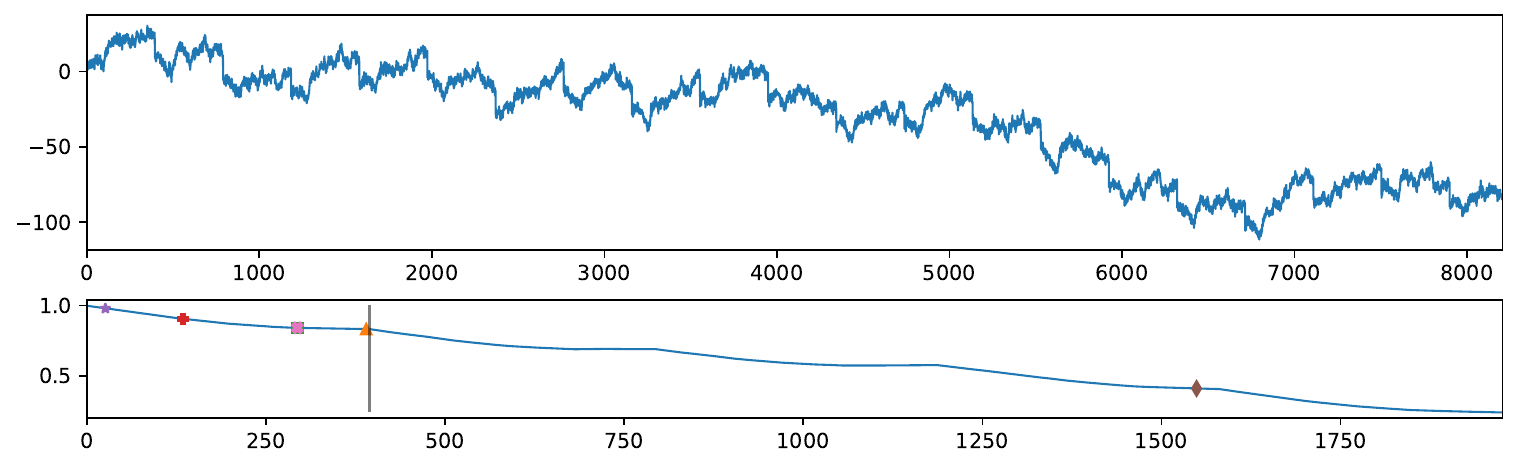}
    \end{subfigure}
    \begin{subfigure}[b]{\columnwidth}
         \centering
         \includegraphics[width=.8\textwidth]{images/window_size_legend}
    \end{subfigure}
    \caption{Three synthetic time series, together with their autocorrelation function. Detection results and ground truth are labeled in the ACF.}
    \label{fig:synthetic_periodicity}
\end{figure}

\begin{table}
    \centering
    \begin{tabular}{lrr}
    \toprule
    Algorithm &  Accuracy &  Avg Runtime (s) \\
    \midrule
    Random     &     0.007 &              - \\
    SuSS~\citep{ermshaus2023clasp} &     0.008 &         0.0113 \\
    MWF~\citep{imani2021multi} &     0.074 &         0.0441 \\
    FFT~\citep{ermshaus2023window}       &     0.169 &         0.0026 \\
    Autoperiod~\citep{vlachos2005periodicity} &     0.512 &         2.6729 \\
    ACF~\citep{ermshaus2023window}        &     0.687 &         0.0006 \\
    Peaks      &     0.754 &         0.0007 \\
    \bottomrule
    \end{tabular}
    \caption{Accuracy and runtime of period detection methods on synthetic datasets.}\label{tab:periodicityresults}

\end{table}

\section{Periodicity Detection Experiments}\label{app:window_size}
We conduct experiments to evaluate periodicity selection on a simple synthetic benchmark. Our signal is composed of three components, a random walk, noise, and a periodic random walk. First we draw a series length $n \propto \text{uniform}(60, 10000)$ and then an integer period $3 < k < \frac{n}{10}$. We draw the noise strength and periodicity strength $\eta_n, \eta_p \propto \text{uniform}(0, 10)$ and then sample a series as 
\[
x_t = w(t) + \eta_n \epsilon(t) + \eta_p w'(t \operatorname{mod} k)
\]
with $w$ and $w'$ random walks and $\epsilon$ independent noise. This very simple mechanism provides us with a ground truth periodicity, sequences of different difficulties (depending on the noise strength, periodicity strength and potential autocorrelation in $w'$). Given our experience with real data, this seems a realistic (if simplified) family of signals, signals that resemble commonly observed patterns, such as heartbeats or weekly traffic patterns. In practice, there is often one dominating periodicity, say daily or weekly patterns in human behavior, where each individual daily or weekly patten can be quite complex (see Figure~\ref{fig:bikesdata}), but is the repetition is quite clear. We also experimented with having more than one periodicity, either with one being a multiple of the other (such as weeks and days) or them being independent. Given the difficulties we encountered with determining a single periodicity, we did not attempt to measure success for multiple periodicities. We limited ourselves to additive seasonal patterns, which we found more challenging than multiplicative patterns; several obvious extensions are possible. Figure~\ref{fig:synthetic_periodicity} shows typical examples of various difficulties; the top shows a simple example, for which all but SuSS find the correct periodicity. The middle figure shows a more challenging series, for which periodicity is easily distinguished by eye on the original series, but on which several methods fail. The bottom shows an very noisy example in which an expert can find the solution based on the ACF, but on which most approaches fail.
Table~\ref{tab:periodicityresults} shows results from the methods investigated in \citet{ermshaus2023window} using their code on 1000 synthetic series. Unfortunately, we were unable to run the RobustPeriod~\citep{wen2021robustperiod} code provided by  \citet{ermshaus2023window}, despite efforts to fix version mismatches. Here, Peaks denotes a heuristic based on the autocorrelation function (ACF) that we investigated that returns the first non-dominated peak i.e. a peak that is higher than any following ACF values. This is only a slight modification of the ACF heuristic proposed by \citet{ermshaus2023window}, who propose to return the first peak. The Random score is calculated by permuting the known responses, i.e. it is  random given the true distribution of periodicities. We conclude that on this synthetic, but realistic and challenging benchmark, current methods are outperformed by simple heuristics, which take much lower runtime. However, room for improvement remains for more sophisticated methods, as all methods still fail on cases where the truth can be determined by visual inspection of the ACF, see Figure~\ref{fig:period_failure}.








\end{document}